\documentclass[runningheads]{llncs}
\usepackage{graphics} 
\usepackage{epsfig} 
\usepackage{mathptmx} 
\usepackage{times} 
\usepackage{amsmath} 
\usepackage{amssymb}  

\usepackage{algorithm}
\usepackage[noend]{algorithmic}
\usepackage{setspace}

\usepackage{caption}

\usepackage{subcaption} 
\usepackage{placeins}

\usepackage{diagbox} 
\usepackage{booktabs} 
\usepackage{xcolor}
\begin{document}
\title{Chemotaxis Based	Virtual Fence for Swarm Robots in Unbounded Environments\thanks{Supported by National Information Technology Development Agency, Nigeria.	
	Simulations were undertaken on ARC3, part of the High Performance Computing facilities at the University of Leeds, UK.}}
%
%

\author{Simon O. Obute\inst{1}
	\and
Mehmet R. Dogar\inst{1} \and
Jordan H. Boyle\inst{2}}
\authorrunning{S. O. Obute et al.}
%
\institute{School of Computing, University of Leeds, UK\\ \email{\{scsoo,M.R.Dogar\}@leeds.ac.uk} \and
Department of Mechanical Engineering, University of Leeds, UK\\
\email{J.H.Boyle@leeds.ac.uk}}
\maketitle              
\begin{abstract}
	This paper presents a novel swarm robotics application of chemotaxis behaviour observed in microorganisms. This approach was used to cause exploration robots to return to a work area around the swarm's nest within a boundless environment. We investigate the performance of our algorithm through extensive simulation studies and hardware validation. Results show that the chemotaxis approach is effective for keeping the swarm close to both stationary and moving nests. Performance comparison of these results with the unrealistic case where a boundary wall was used to keep the swarm within a target search area showed that our chemotaxis approach produced competitive results.
	\keywords Chemotaxis \and Swarm robots \and Exploration \and Distributed robot systems
\end{abstract}

\section{Introduction}\vspace{-.2cm}
Swarm robotics is a bio-inspired multi-robot research theme focused on the actualization of swarm intelligence observed in nature on robotic platforms. Biological swarms like bees, ants and termites are able to accomplish complex tasks, such as finding food and building nests, through local interaction with each other and/or their environments. These tasks are beyond the capabilities of a single agent and, in general, unattainable without cooperation among swarm members. By mimicking nature, swarm robotics emphasizes local interactions and autonomous decision making of agents to develop simple, flexible, scalable and robust algorithms for multi-robot platforms \cite{Bayindir2016,Zedadra2015}.
Typical swarm behaviours include foraging, aggregation, exploration, clustering, assembly and flocking \cite{Bayindir2016}.
A major concern for swarm robotics applications is development of effective means for keeping swarm robots within the desired work area while they perform their tasks. This is important for real world deployment of swarm robotics systems, where they encounter unknown and unstructured environments that are, in many cases, unfenced.  Much work has been done that assumed the presence of a boundary (or fence) \cite{Arvin2018,Lima2017a,Schmickl2011} that keeps the swarm from drifting over time from the work area - an approach we believe to be unrealistic because such a structure will not be available in many applications and  in some cases the swarm working area must change over time. Thus, a main contribution of this work is to provide a means for keeping swarm robots within a work area by introducing a simple, hardware-grounded means of communication between robots and their nest (or guide robot). Our approach is effective for both static and dynamic work area for swarm robots. We make use of a nest robot that broadcasts a range-limited signal that degrades with distance, which swarm members listen to.
When they sense that intensity of nest signal drops below a threshold, they use this sensory information to perform a chemotaxis-based search for the work area.

This simple bio-inspired search algorithm is based on the chemotaxis behaviour observed in the nematode worm \textit{Caenorhabditis elegans} \cite{pierce2005analysis,Ward1973} in response to chemical attractants, which is one of the primary methods the worm uses to navigate towards favourable conditions. The worm's small size and limited neural circuit preclude the use of `stereo' sensing to detect the spatial gradient of the chemical cue, so the behaviour is instead based on the temporal gradient sensed by the worm as it moves, which requires only a single sensory receptor. By default the worm performs a random walk consisting of runs of relatively straight, forwards motion, interspersed by large  turns called pirouettes at random intervals. If the temporal gradient of a chemical attractant is positive, the probability of performing a turn is reduced so the worm is more likely to keep moving in a beneficial direction. Conversely, a negative temporal gradient increases the turn probability so it is more likely to reorient to a more favourable direction.

The results we present in this paper study the relationship between nest signal threshold and work area size, selection of good chemotaxis parameters for stationary and moving nests, and how nest velocity affects target search efficiency of swarms.

Section \ref{sec_literature_review} reviews swarm algorithms in literature, with a focus on mechanisms used to keep robots within work area. In Section \ref{sec_unbounded_exploration} we detail the chemotaxis-based algorithm for keeping robots within a designated work area for both stationary and moving nest (or guide robot). The details of how we model our swarm communication is presented in Section \ref{sec_swarm_comm_model}. In Sections \ref{sec_simulation_results} and \ref{sec_robot_experiments} we present simulation and real robot experimental results respectively, then conclude in Section \ref{sec_conclusion}.

\vspace{-.2cm}
\section{Review of Similar Works}\label{sec_literature_review}\vspace{-.2cm}
The two extremes in multi-robot exploration algorithms are: random search and systematic exploration\cite{Trianni2015}. In random search, robots use Brownian-like motion to explore the environment until they perceive a feature of interest. This approach is sufficient for bounded environments because it will typically explore all regions of the environment when given sufficient time. It is unsuitable for large or open (unbounded) environments because robots will drift away from the work area and lose contact with other swarm members.
In the systematic approach, robots use a priori knowledge of the environment's structure to methodically explore it. Although this approach optimizes exploration time and prevents oversampling of regions in the environment, its memory requirements become excessive for large environments. Its localization, mapping and planning algorithms do not scale well with increase in swarm and environment sizes.
Most swarm robotics exploration algorithms propose balance between these two approaches to develop robust, flexible and scalable algorithms.

The Gradient Algorithm in \cite{Hoff2012} is based on the gas expansion approach, where robots try to maintain communication links with their neighbours while maximizing the distance between themselves. Also in \cite{Hoff2012}, the Sweeper Algorithm made the interconnected robots form a 1D chain, which rotates about the nest (like the hand of the clock, where the nest is the centre). This extends the area covered by the swarm  beyond what is attainable by the Gradient Algorithm. In these algorithms, the size of environment the swarm can cover is dependent on the number of robots, since maintaining communication links is paramount for the swarm to keep robots within work area. The success of these approaches also requires formation of \emph{ad hoc} networks where such infrastructure is unavailable \cite{Winfield2008}. The work in \cite{Couceiro2014a} and hierarchical swarm in \cite{Ngo2014} also require the swarm to maintain communication networks for their task. In \cite{Tolba2017}, swarms of robots deployed underwater kept track of their initial deployment region using a scheme termed virtual tether search. The robots used random walk to search for targets, while using dead reckoning to constrain their distance from their initial deployment point. Another dead reckoning approach, which kept track of a stationary nest location within the context of foraging swarm was implemented in \cite{Lu2018}. Dead reckoning is unsuited for large work areas because it becomes less accurate over time or distance travelled, and is terrain dependent. It is also unusable for applications where the nest is mobile. Pheromone-based approaches, as in \cite{Lima2017a} and \cite{Zedadra2015} are difficult to realise in hardware. The various attempts to provide hardware implementations have resulted in the use of beacon robots \cite{Hoff2012}, LCD platform \cite{Arvin2018}, RFID and other technologies \cite{Zedadra2017}. Such approaches do not scale well when increasing size of the environment.

Our approach does not require network connectivity among swarm robots or dead reckoning, thus freeing them to autonomously explore the work area. Our robots only need to sense the intensity of a nest signal (we use sound in the present work) in their current location to make autonomous decision on whether they are within or outside the desired work area. This approach greatly simplifies our swarm algorithm and communication strategy. Furthermore, we demonstrate that our implementation is effective for a moving nest and easily realizable on hardware platforms.

\vspace{-.2cm}
\section{Unbounded Exploration}\label{sec_unbounded_exploration}\vspace{-.2cm}
\subsection{Robot Exploration with Chemotaxis Activation}
\begin{figure}
	\centering
\begin{minipage}[!t]{.55\textwidth}
	\begin{algorithm}[H]
		\centering
			\begin{algorithmic}[1]
				\STATE Sense nest signal, $A_{t}$
				\STATE Initialize $P_{t} = P_{b}$ to default value
				\IF{$A_{t} < A(d_{c})$}
					\IF{$A_{t} < A_{t-1}$}
						\STATE 	$P_{t} = P_{b} \times M$
					\ELSIF{$A_{t} > A_{t-1}$}
						\STATE $P_{t} = P_{b} \div D$
					\ENDIF
				\ENDIF
				\IF{rand(0,1) $ < P_{t}$}
					\STATE make random turn of $\mathcal{N}(180^{0},90^{0})$
				\ELSE
					\STATE make straight motion
				\ENDIF
			\end{algorithmic}
	\caption{Random Walk with chemotaxis activation.}
	\label{alg_rw_chemotaxis}
	\end{algorithm}
\end{minipage}\hfill
\begin{minipage}[h]{0.34\textwidth}\vspace{1cm}
		\begin{subfigure}[t]{\textwidth}
			\centering
			\scalebox{1}[1]{\includegraphics[width=0.45\textwidth]{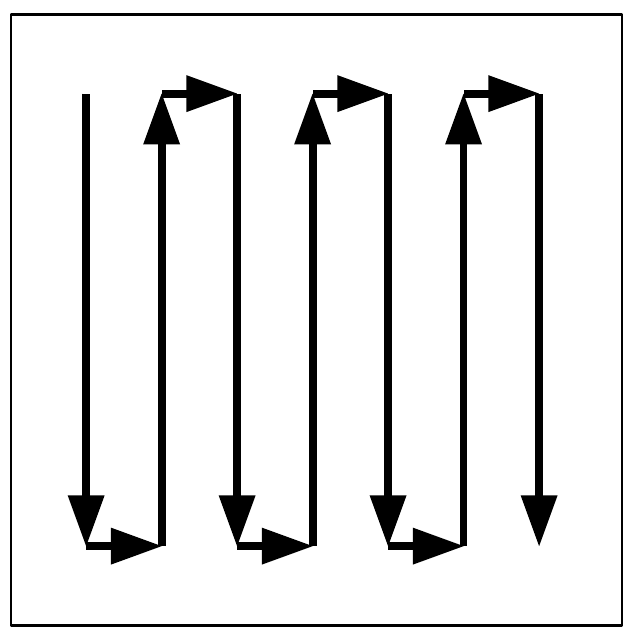}}
			\subcaption{Vertical.}
			\label{fig_nest-navigation-x}
		\end{subfigure}
	
		\begin{subfigure}[t]{\textwidth}
			\centering
			\scalebox{-1}[1]{\includegraphics[width=0.45\textwidth,angle=90]{./figures/nest-navigation}}
			\subcaption{Horizontal.}
			\label{fig_nest-navigation-y}
		\end{subfigure}
		\caption{Nest search behaviour.}
		\label{fig_nest-navigation}
\end{minipage}
\vspace{-.5cm}
\end{figure}

\vspace{-.2cm}
In our design, the region surrounding the nest (or guide robot) is divided into a work area ($< d_{c}$) and a chemotactic region ($>= d_{c}$), where $d_{c}$ is the distance that corresponds to chemotaxis activation threshold, $A(d_{c})$. Within the work area, the robots perform their expected swarm task, which we abstract as random exploration of the region. Beyond this area, there is the chemotactic region, which serve as an effective wall for keeping robots within the work area. Robots within the chemotactic region make use of a \textit{C. elegans}-inpsired `chemotaxis' behaviour (using sound intensity in place of a chemical signal) to search for the work area. Algorithm \ref{alg_rw_chemotaxis} represents the steps executed by each exploration robot in the swarm within each time step.

The robot first senses the nest signal, $A_{t}$ from its current location and initializes its turn probability, $P_{t}$, to a pre-determined base probability, $P_{b}$. If $A_{t}$ is less than a pre-determined threshold, $A(d_{c})$, it updates $P_{t}$ based on whether nest signal has increased or decreased since the last time step. $M$ and $D$ are probability multipliers and divisors for increasing or decreasing the robot's turn probability. The robot uses $P_{t}$ to decide whether to make a turn or continue linear, straight motion at constant velocity, $v_{r}$. 
\subsection{Moving the Nest}\vspace{-.2cm}
In the basic form of our approach, the nest is stationary. However, with the absence of a physical boundary to restrict the swarm's work area, the nest itself can be free to move within the unbounded environment, thereby guiding exploration of the swarm as it makes its motion. The basic form of this moving nest is a linear motion from a starting location to destination point by moving at a constant velocity that is a fraction of the exploration robots' velocity $v_{n}$. The nest waits (stops briefly) whenever it senses robot(s) within $d_{n}$ metres of its front region to avoid collisions.
We extend the nest's motion to cover a 2D search area by following a sequence of checkpoints that causes it to perform vertical  then horizontal sweeps of the environment (as shown in Fig. \ref{fig_nest-navigation}). These sweeps are repeated continuously for a maximum simulation time, $t_{max}$.
The introduction of a moving nest extends the search area of the swarm of robots by guiding them to regions where they can execute their tasks. However, it also brings up questions regarding the optimal nest velocity that will give the best balance between exploration speed, accuracy (or efficiency) and minimization of the number of robots that lose track of the nest signal and get left behind. We investigate these questions in upcoming sections.

\vspace{-.2cm}
\section{Development of Communication Model}\label{sec_swarm_comm_model}\vspace{-.2cm}
The successful deployment of our swarm in unbounded environments is dependent on a realistic communication model between the nest (or guide) and other swarm members. We implemented this communication using white noise broadcast from a speaker attached to the nest. To model the noise accurately in our simulations, we first collected sound intensity data from real robot experiments. In the setup, a Turtlebot2 robot was placed 15m away from the speaker and programmed to move towards the sound source at a velocity of 0.1 m/s while logging the sound intensity perceived on its attached microphone. This experiment was repeated 5 times.

In the second step, we computed parameters for sound degradation with distance. We used Equation \ref{eqn_sound_eqn} for the sound model \cite{Yu2017}, where $A(d)$ is the sound intensity $d$ metres away from a speaker. $A_{0}$ is the sound intensity at the speaker, $\alpha$ is the attenuation factor and the $A_{e}$ term was added to account for ambient noise. The model parameters were computed by evaluating the least square error fit between the collected data and Equation \ref{eqn_sound_eqn} using MATLAB's nonlinear curve-fitting function. The values computed were $A_{0} = 140.5193$, $\alpha = 0.1193$ and $A_{e} = 48.1824$.

\begin{equation}
A(d) = A_{0}e^{-\alpha d} + A_{e}\label{eqn_sound_eqn}
\end{equation}

\begin{figure}[t]
	\centering
	\begin{subfigure}{0.47\textwidth}
		\includegraphics[width=\textwidth]{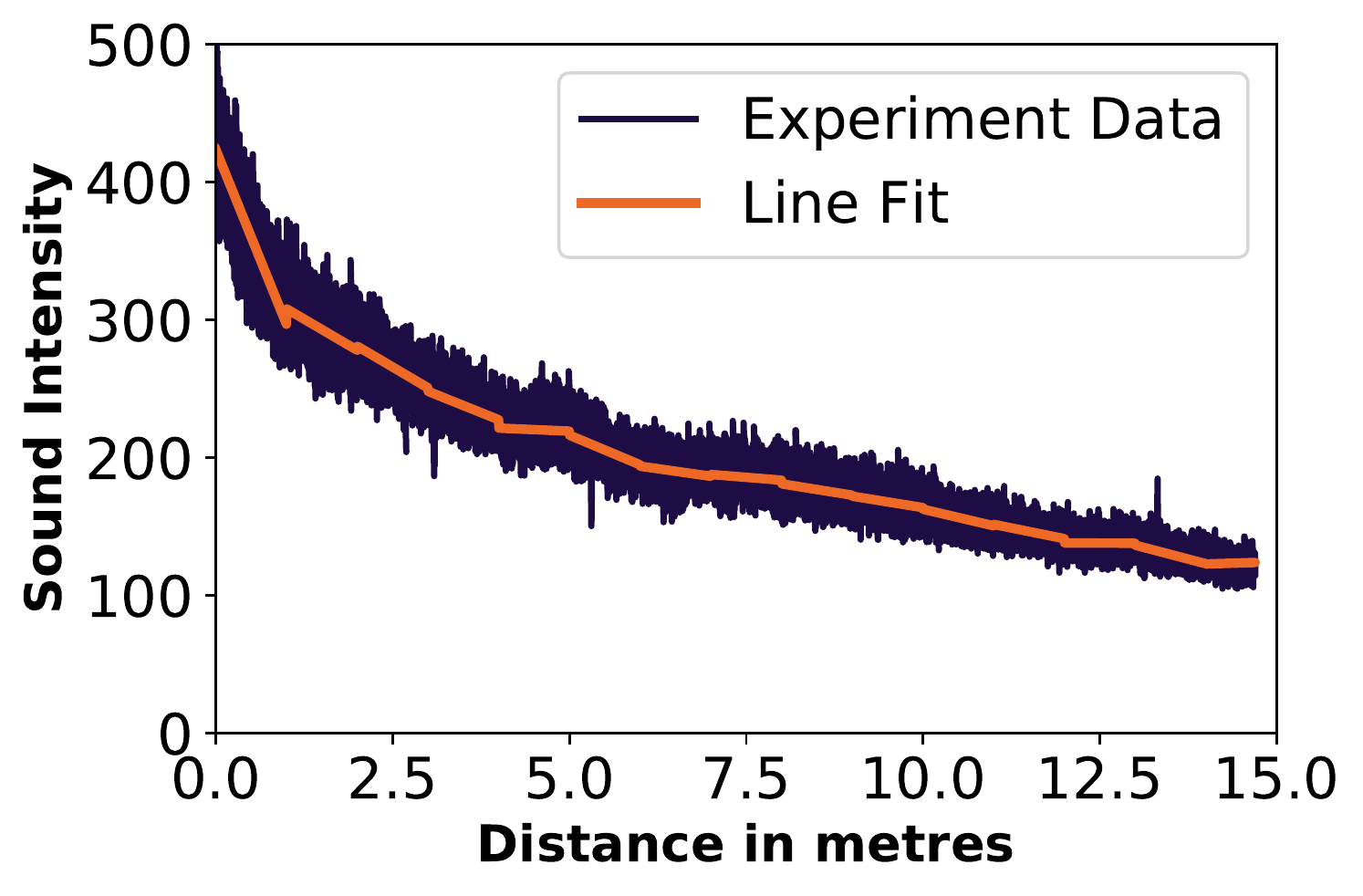}
		\subcaption{Experiment Data represents overlayed raw data from 5 experiments.}
		  
		\label{fig_line-fits-noise-model}
		
	\end{subfigure}\hfill
	\begin{subfigure}{0.47\textwidth}
		\includegraphics[width=\textwidth]{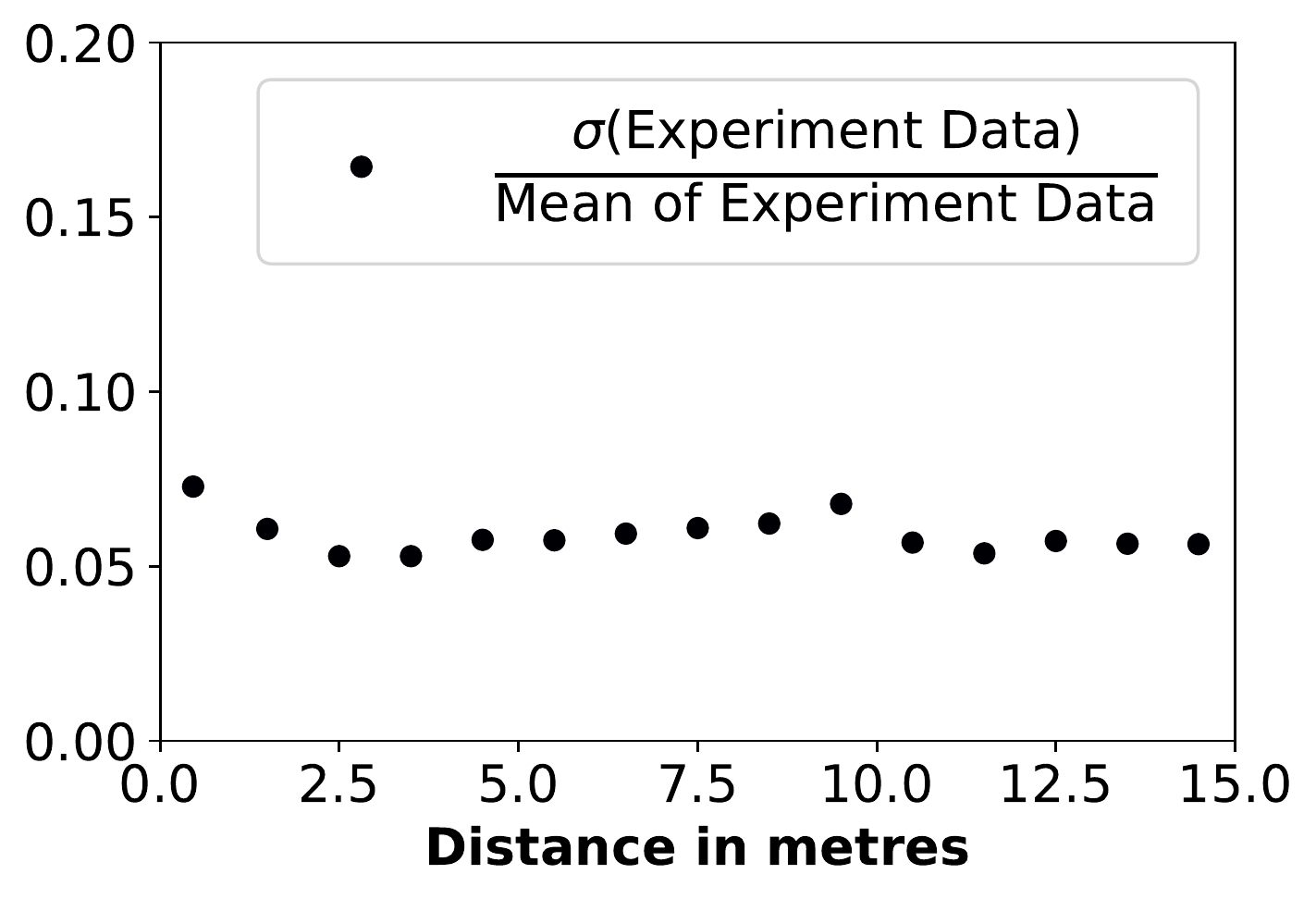}
		\subcaption{Ratio of standard deviation of experiment data to mean value of line fit segment.}
		\label{fig_noise-std-mean-plot}
		
	\end{subfigure}
	\caption{Developing noise model for simulation.}
	\label{fig_noise_model_development}\vspace{-.5cm}
\end{figure}

The third step involved quantifying the noise in the recorded data. To do this, the logged data was broken into 1 metre segments; a line was fitted to each segment (as shown in Fig. \ref{fig_line-fits-noise-model}); and the means and standard deviations of the sound data from the fitted line segments were computed. It was observed that, though the deviation increased with increasing mean sound intensity, the ratio between each mean and the corresponding standard deviation remained fairly constant at 0.06 (Fig. \ref{fig_noise-std-mean-plot}). Thus, noise was modelled as random deviation with standard deviation of 0.06 from the mean intensity perceived from a sound source. Equation \ref{eqn_noisy_sound} represents the noisy sound intensity $d$ metres away, at the $i$-th time step.

\begin{equation}
	A_{i}(d) = A(d)\left(1-\mathcal{N}(0,0.06)\right)\label{eqn_noisy_sound}
\end{equation}

In order to help the robots detect the underlying gradient despite the substantial noise, we added a filtering system to the behavioural algorithm. We implemented an averaging filter, shown in Equation \ref{eqn_average_filter}, where $t$ is the current time step and $n = 40$ time steps. A time step in our experiments was 0.0025 second, which means that a robot's sensed nest signal is updated at 1Hz (40 time steps make 1 second).
\begin{equation}
 A_{f}(d) = \frac{\sum_{i=t-n}^{t}A_{i}(d)}{n}\label{eqn_average_filter}
\end{equation}

\vspace{-.2cm}
\section{Simulation Results}\label{sec_simulation_results}\vspace{-.2cm}
\subsection{Simulation Setup}\vspace{-.2cm}
We used the Gazebo simulation platform to investigate the performance of our approach. A swarm size of 10 robots moving at $v_{r} = 0.605$ m/s was used. The robot's base turn probability for all simulations was $P_{b} = 0.0025$ per time step. Nest velocity $v_{n}$ is expressed relative to $v_{r}$ for all experiments and the nest stops when it senses a robot is within $d_{n} = 0.1$m from its front region.

\begin{table}[t]
	\centering
	\caption{Chemotaxis activation distance $d_{c}$ versus robots distances from nest $d_{r}$. Each value represents the mean number of robots within $d_{r}$ of the nest and the 95\% confidence interval of this value over 30 repetitions of each simulation. Simulation time was 1500 seconds, $M = 10$ and $D = 1000$.}\label{table_threshold_vs_robots_in_range}
	\resizebox{0.65\textwidth}{!}{
		\begin{tabular}{l@{\hskip 12pt}l@{\hskip 12pt}l@{\hskip 12pt}l@{\hskip 12pt}l@{\hskip 12pt}l}
			\toprule
			\diagbox[width=3em]{$\boldsymbol{d_{r}}$}{$\boldsymbol{d_{c}}$} &               6m  &               8m  &               10m &               12m &               14m \\
			\midrule
			6  &   $4.4 \pm 0.03$ &         -         &   	-			&   -				&   	-			 \\
			8  &   $7.9 \pm 0.02$ &   $5.3 \pm 0.04$ &   			-   &   -				&   -				 \\
			10 &   $9.4 \pm 0.01$ &   $8.3 \pm 0.02$ &   $6.0 \pm 0.04$ &   -				&   -				 \\
			12 &   $9.8 \pm 0.01$ &   $9.5 \pm 0.01$ &   $8.6 \pm 0.02$ &   $6.5 \pm 0.04$ &   -				 \\
			14 &  $10.0 \pm 0.00$ &   $9.8 \pm 0.01$ &   $9.6 \pm 0.01$ &   $8.8 \pm 0.02$ &   $6.7 \pm 0.04$ \\
			16 &  $10.0 \pm 0.00$ &  $10.0 \pm 0.00$ &   $9.9 \pm 0.01$ &   $9.7 \pm 0.01$ &   $8.7 \pm 0.03$ \\
			18 &  $10.0 \pm 0.00$ &  $10.0 \pm 0.00$ &  $10.0 \pm 0.00$ &   $9.9 \pm 0.00$ &   $9.6 \pm 0.01$ \\
			20 &  $10.0 \pm 0.00$ &  $10.0 \pm 0.00$ &  $10.0 \pm 0.00$ &  $10.0 \pm 0.00$ &   $9.9 \pm 0.01$ \\
			22 &  $10.0 \pm 0.00$ &  $10.0 \pm 0.00$ &  $10.0 \pm 0.00$ &  $10.0 \pm 0.00$ &  $10.0 \pm 0.00$ \\
			
			\bottomrule
	\end{tabular}}\vspace{-.2cm}
\end{table}

\vspace{-.2cm}
\subsection{Determining the Chemotactic Region}\vspace{-.2cm}
To be able to determine the chemotaxis activation intensity, $A(d_{c})$, needed to keep swarm robots within a specified distance of a stationary nest, $d_{r}$, we conducted simulation experiments where chemotaxis activation distance, $d_{c}$ was varied from 6m to 14m from the nest. The robots were made to perform a random walk for 1500 seconds around the nest for each $d_{c}$. Each simulation was repeated 30 times and we analysed the average number of robots within varied distances from the nest. Results are presented in Table \ref{table_threshold_vs_robots_in_range}. This shows that the chemotactic region is effective in keeping more than 95\% of robots within $d_{c} + 4$ of the nest in all distance ranges tested. Thus, for design purposes, to keep at least 95\% of swarm members within $d_{w}$ metres of a stationary nest, the chemotaxis activation distance can be computed using Equation \ref{eqn_dc_vs_dw}. 
\begin{equation}
	d_{c} = d_{w} - 4
	\label{eqn_dc_vs_dw}
\end{equation}

\subsection{Effects of Base Probability Multiplier and Divisor}
\begin{figure}[t]
	\centering
	\begin{subfigure}[b]{0.3\textwidth}
		\centering
		\scalebox{1}[1]{\includegraphics[width=.95\textwidth]{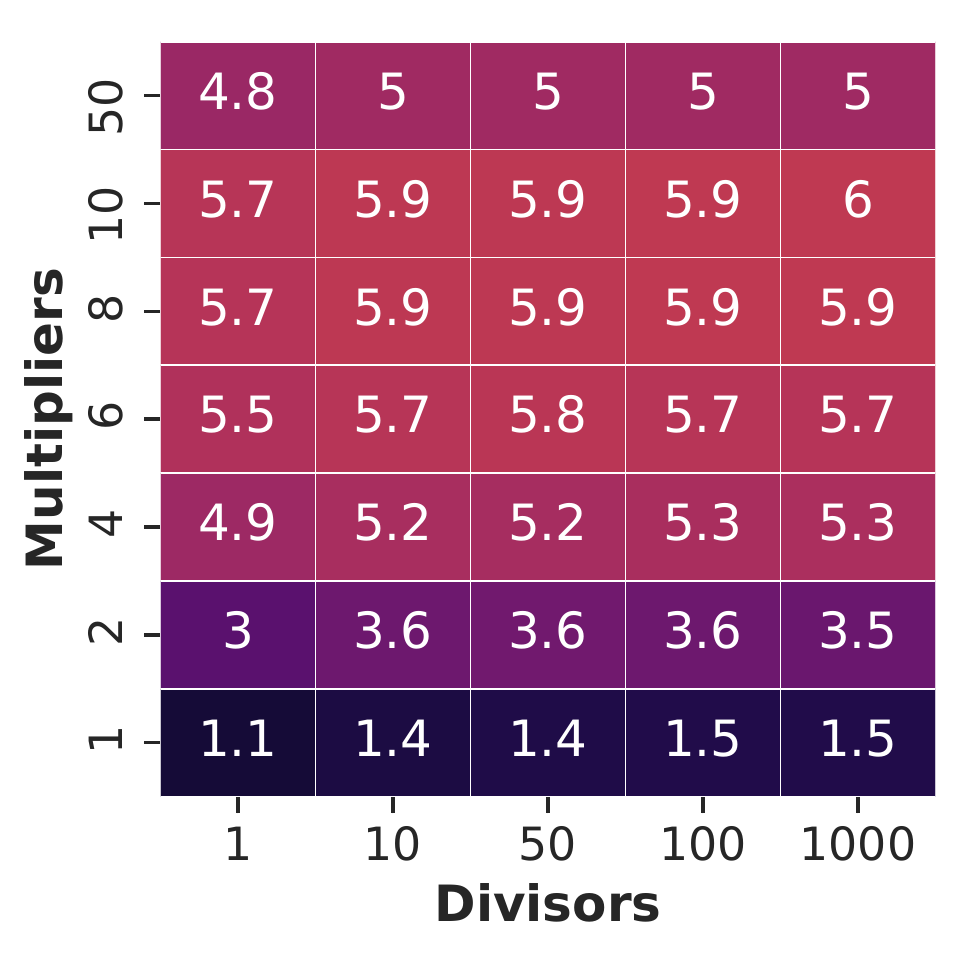}}\vspace{-.2cm}
		\subcaption{$v_{n} = 0$, $d_{r} = 10$m.}
		\label{fig_NA0-10}
		\scalebox{1}[1]{\includegraphics[width=.95\textwidth]{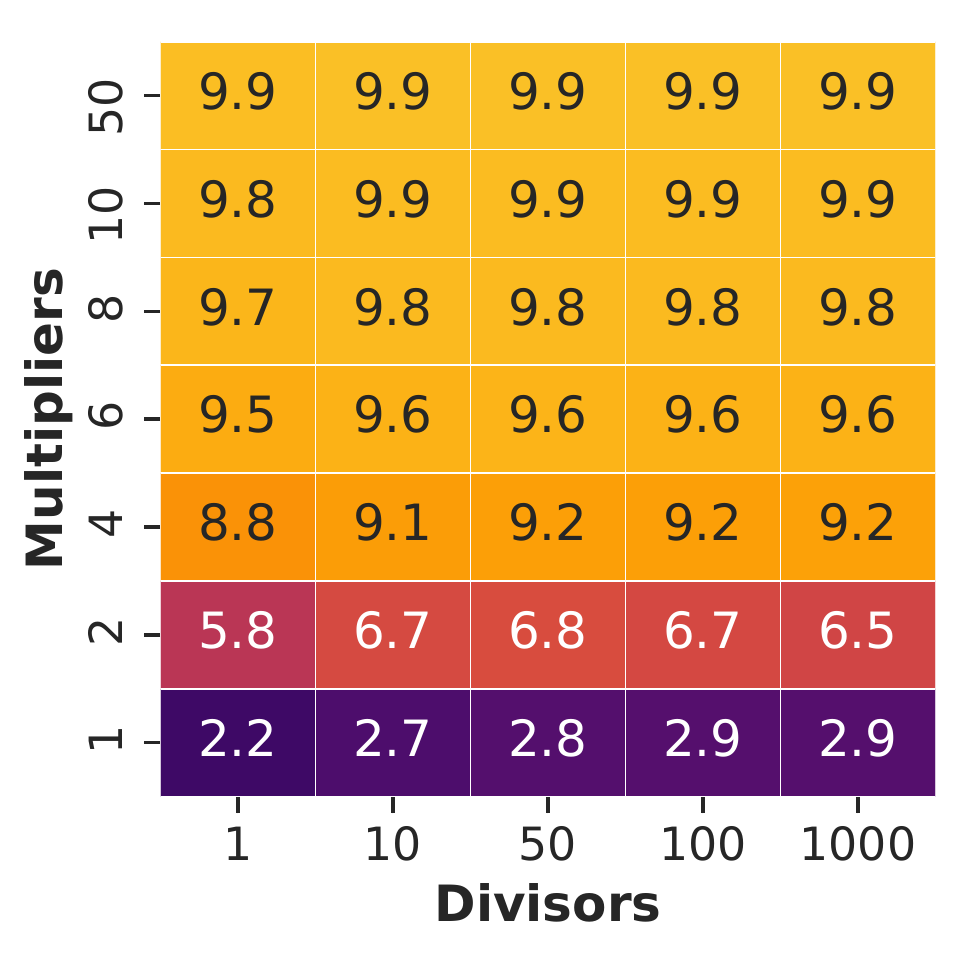}}\vspace{-.2cm}
		\subcaption{$v_{n} = 0$, $d_{r} = 16$m.}
		\label{fig_NA0-16}
	\end{subfigure}\hfill
	\begin{subfigure}[b]{0.3\textwidth}
		\centering
		\scalebox{1}[1]{\includegraphics[width=.95\textwidth]{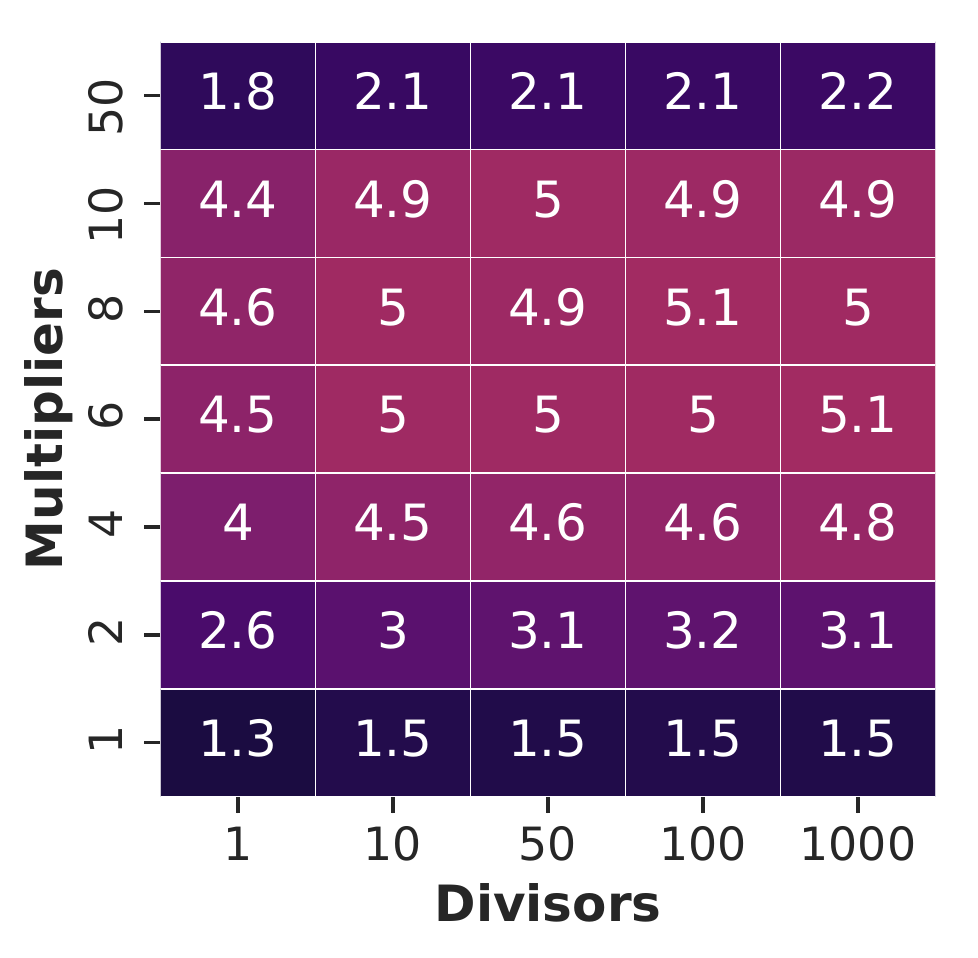}}\vspace{-.2cm}
		\subcaption{$v_{n} = 0.125$, $d_{r} = 10$m.}
		\label{fig_NA0p125-10}
		\scalebox{1}[1]{\includegraphics[width=.95\textwidth]{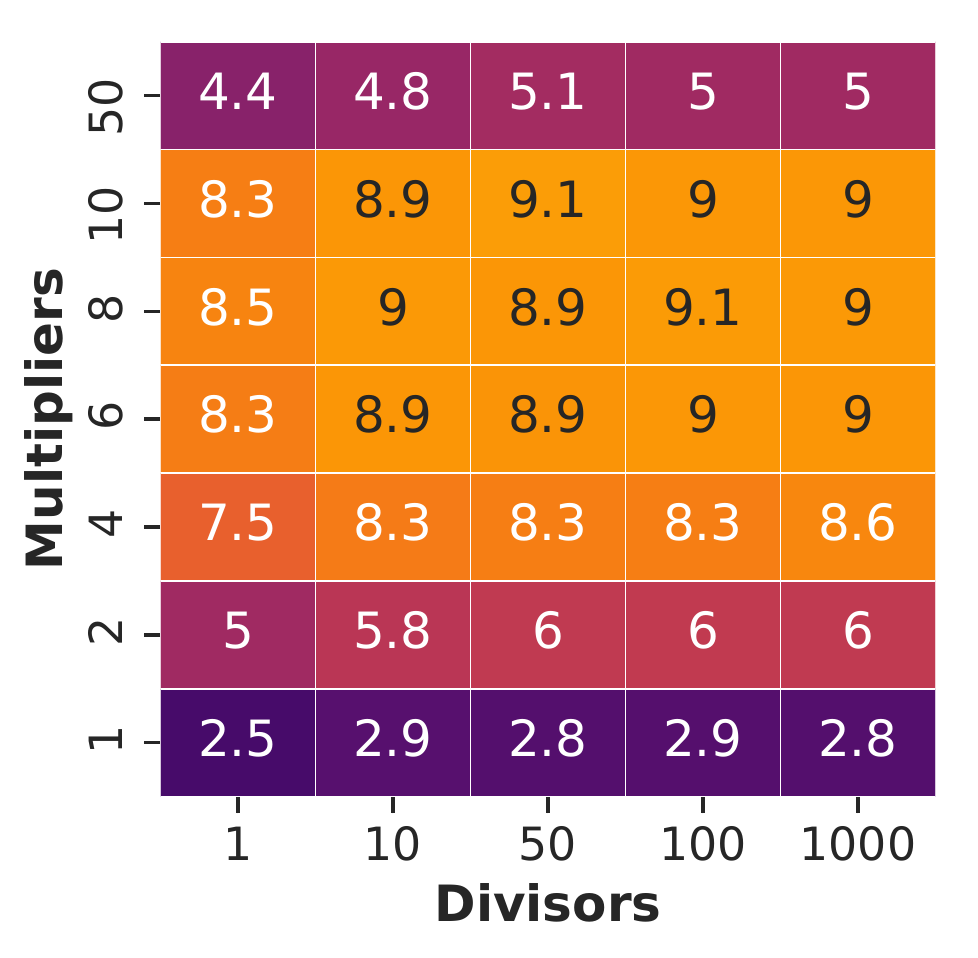}}\vspace{-.2cm}
		\subcaption{$v_{n} = 0.125$, $d_{r} = 16$m.}
		\label{fig_NA0p125-16}
	\end{subfigure}\hfill
	\begin{subfigure}[b]{0.3\textwidth}
		\centering
		\scalebox{1}[1]{\includegraphics[width=.95\textwidth]{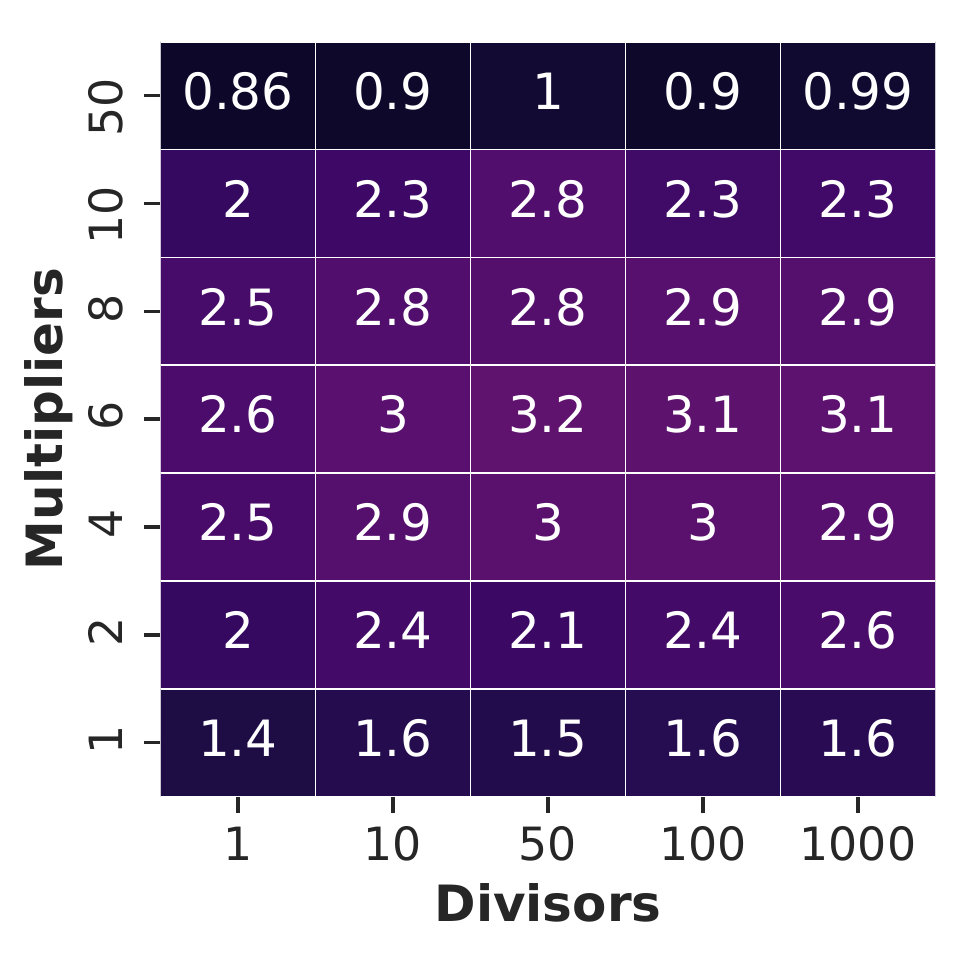}}\vspace{-.2cm}
		\subcaption{$v_{n} = 0.25$, $d_{r} = 10$m.}
		\label{fig_NA0p25-10}
		\scalebox{1}[1]{\includegraphics[width=.95\textwidth]{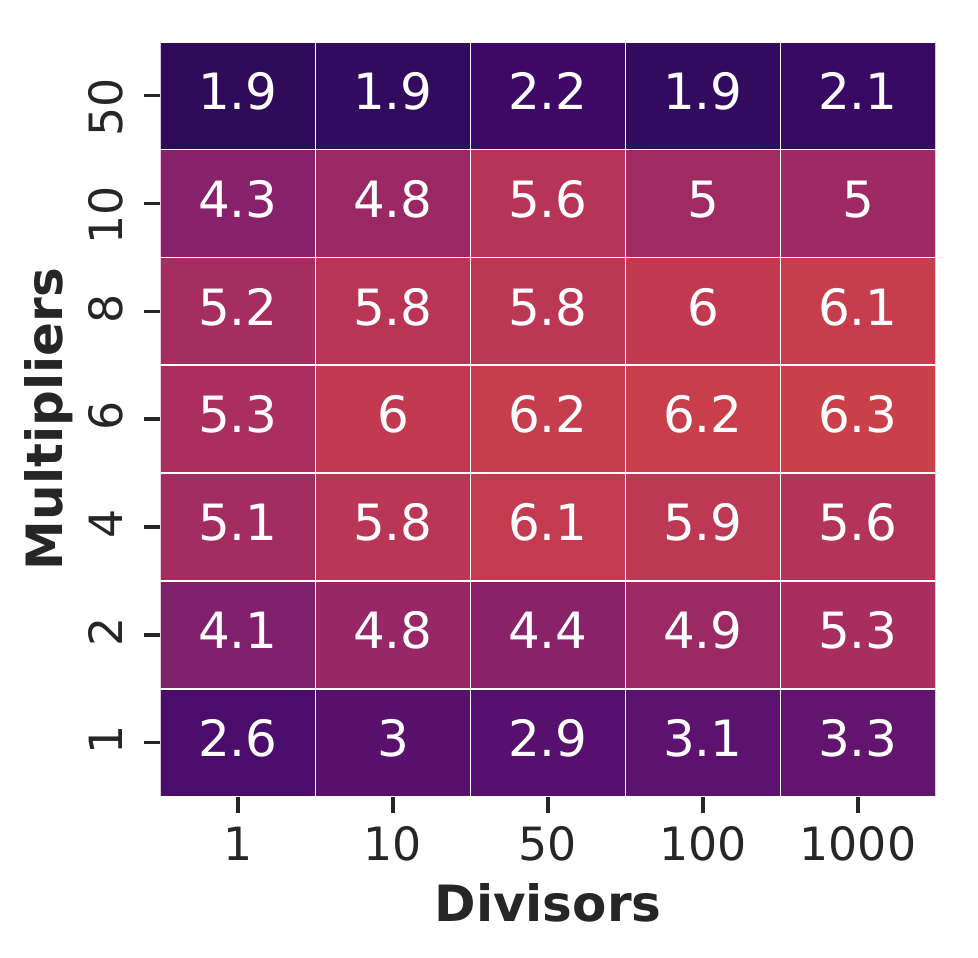}}\vspace{-.2cm}
		\subcaption{$v_{n} = 0.25$, $d_{r} = 16$m.}
		\label{fig_NA0p25-16}
	\end{subfigure}
	\caption{Heat map of the effects of nest relative velocity, $v_{n}$, probability multiplier, $M$, and divisor, $D$, on the average number of robots within a specific distance, $d_{r}$ from the nest/guide robot. Chemotaxis activation distance, $d_{c} = 10$m, for all simulations. 
	}
	\label{fig_NA-vel-M-D}\vspace{-.4cm}
\end{figure}

\vspace{-.2cm}
Important factors that change the effectiveness of the chemotactic region in keeping the swarm together include the nest's relative velocity $v_{n}$, and the probability multiplier $M$ and divisor $D$. We investigate the effects of these values in Fig. \ref{fig_NA-vel-M-D}. The chemotaxis activation distance was 10m for these simulations. Each value in Fig. \ref{fig_NA-vel-M-D} represents the mean number of robots within a distance $d_{r}$ of the nest, averaged across 30 independent simulation repetitions. When $v_{n} > 0$, the nest moved for 100m along a straight path.

The results in Fig. \ref{fig_NA-vel-M-D} indicate that the probability multiplier, $M$, and divisor $D$ play a major role in the effectiveness of using chemotaxis to keep robots within the work area. A small value of $M$ made the robots less responsive to decreasing nest signal when in the chemotactic region, thus causing them to move further away from the work area, indicating a high flexibility of the `chemotactic wall'.  Large values for $M$ made the robots more responsive to negative temporal gradients of the nest signal, preventing them from going further into the chemotactic region. However, a very high probability multiplier, $M = 50$, caused the robots to turn too frequently in the chemotactic region, thus, preventing them from making sufficient linear motion to compute a reliable temporal gradient from the noisy nest signal.

The probability divisor, $D$, has a lesser effect than $M$ on the swarm's ability to remain within the work area. The results show a general trend, where increasing $D$ results in slightly more robots remaining within the work area. Low values of $D$  made robots less responsive to positive temporal gradients of nest signal when in the chemotactic region, making robots' suppression of turns less effective during chemotaxis. Increasing $D$ caused robots performing chemotaxis to suppress turns better when they sense positive temporal gradient from the nest signal.

Fig. \ref{fig_NA-vel-M-D} also shows that fewer robots are able to remain within the work area as the nest's relative velocity, $v_{n}$, increases, which is unsurprising. However, there is a variation in the best performing $M$ and $D$ as the nest's velocity increases. In general, as the nest becomes faster, smaller $M$ and larger $D$ gave better performance. Thus, when $v_{n} = 0$, $M = 10$ performed best, while when $v_{n} = 0.25$, $M = 6$ and $D = 1000$ gave good results. These indicate that an equivalent of Equation \ref{eqn_dc_vs_dw} for the moving nest case is more complex, needing a relationship that relates nest velocity, rate of robots getting `lost' and distribution of robots within the work area.

\vspace{-.2cm}
\subsection{Investigating Exploration Effectiveness}
\vspace{-.2cm}
It is important that the swarm effectively explores the work area, and are able to minimize time spent in the chemotactic region. We tested the swarm's ability to explore the work area using a target search task for both the stationary and moving nest setups. In each simulation, the task is for 10 swarm robots to locate 100 targets that are randomly but uniformly distributed within the search area. Robots were able to detect targets beneath them i.e. when the robot's distance from the target was less than robot's radius. Detected targets were removed from the world and 30 independent simulations were repeated for each simulation setup. Our approach was compared with two environment setups as baseline: when a wall was used to keep robots within search area (\emph{Bounded}); and when the wall was removed (\emph{Unbounded}), thus removing any mechanism to restrict the robots to the search area.

\begin{figure*}[t]
	\centering
	\begin{subfigure}[b]{0.22\textwidth}
		\centering	
		\scalebox{1}[1]{\includegraphics[width=\textwidth]{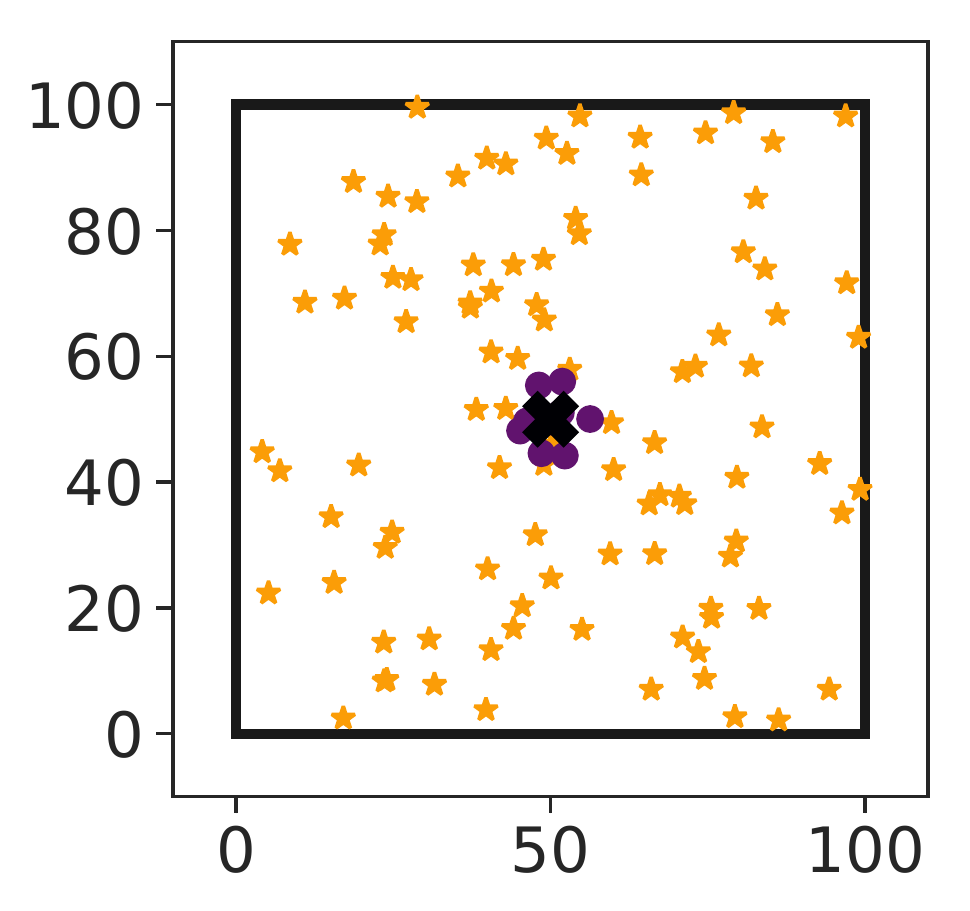}}
		\subcaption{10 seconds.}	
		\label{fig_NA0-AtD0-M1-D1_9p175}
	\end{subfigure}\hfill
	\begin{subfigure}[b]{0.22\textwidth}
		\centering
		\scalebox{1}[1]{\includegraphics[width=\textwidth]{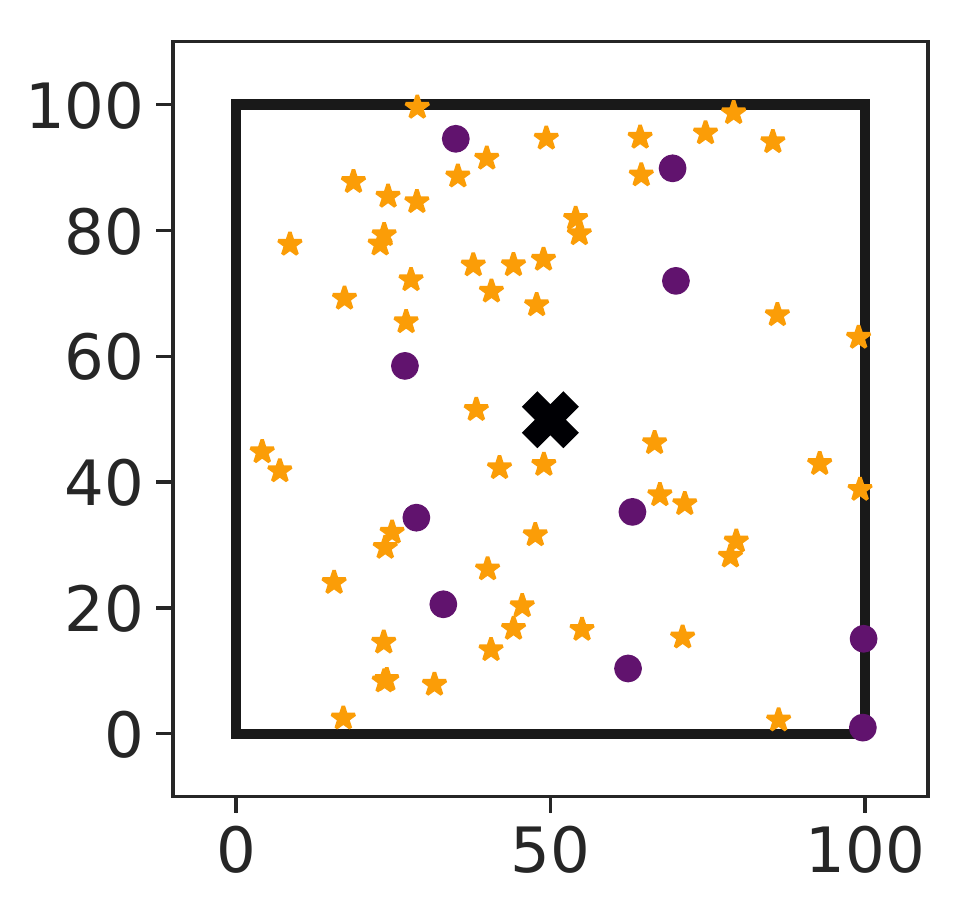}}
		\subcaption{3,500 seconds.}
		\label{fig_NA0-AtD0-M1-D1_3367p175}	
	\end{subfigure}\hfill
	\begin{subfigure}[b]{0.22\textwidth}
		\centering
		\scalebox{1}[1]{\includegraphics[width=\textwidth]{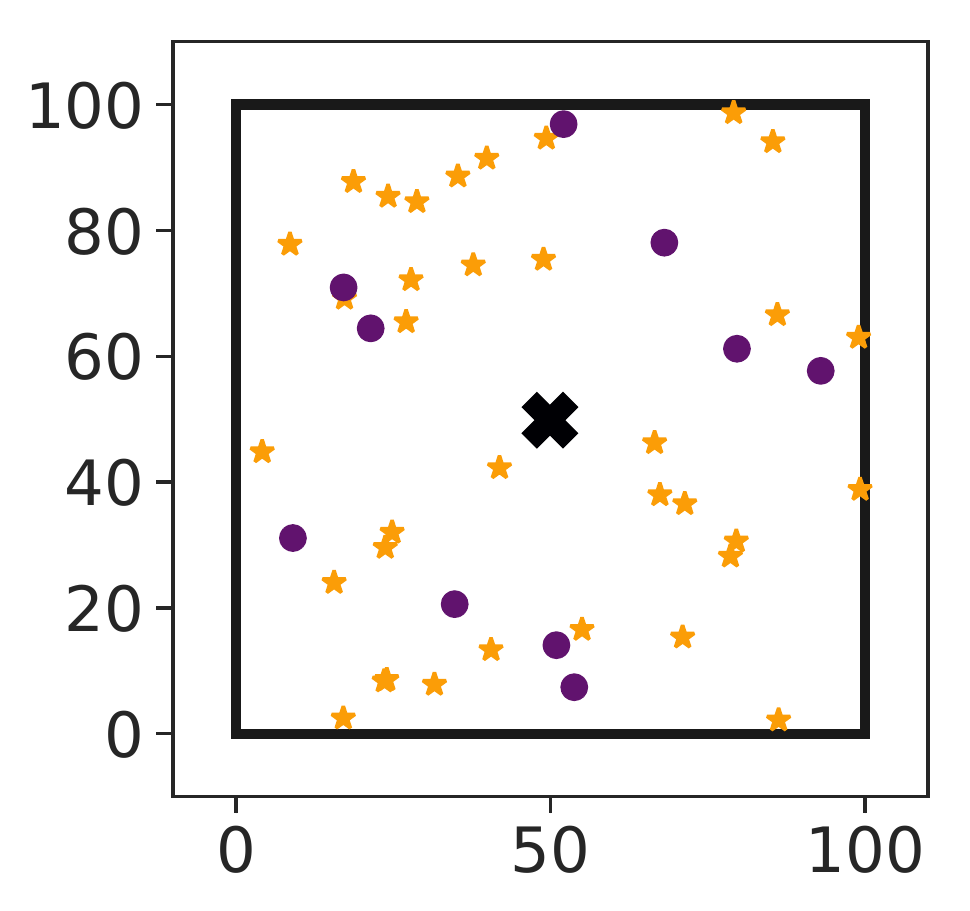}}
		\subcaption{7,000 seconds.}
		\label{fig_NA0-AtD0-M1-D1_6725p175}
	\end{subfigure}\hfill
	\begin{subfigure}[b]{0.34\textwidth}
		\centering
		\scalebox{1}[1]{\includegraphics[width=\textwidth]{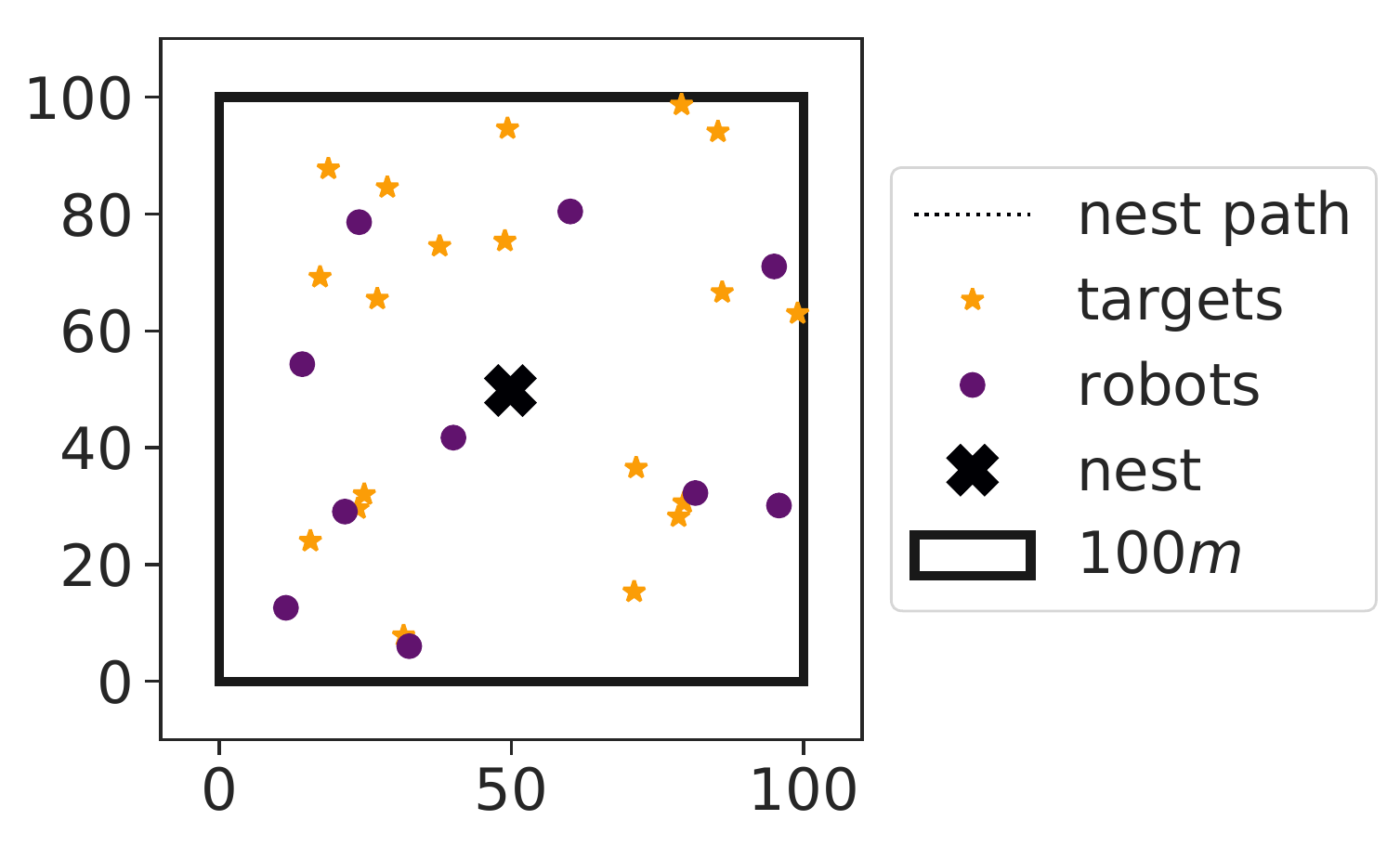}}
		\subcaption{10,000 seconds.}
		\label{fig_NA0-AtD0-M1-D1_10092p375}	
	\end{subfigure}
	
	\begin{subfigure}[b]{0.22\textwidth}
		\centering
		\scalebox{1}[1]{\includegraphics[width=\textwidth]{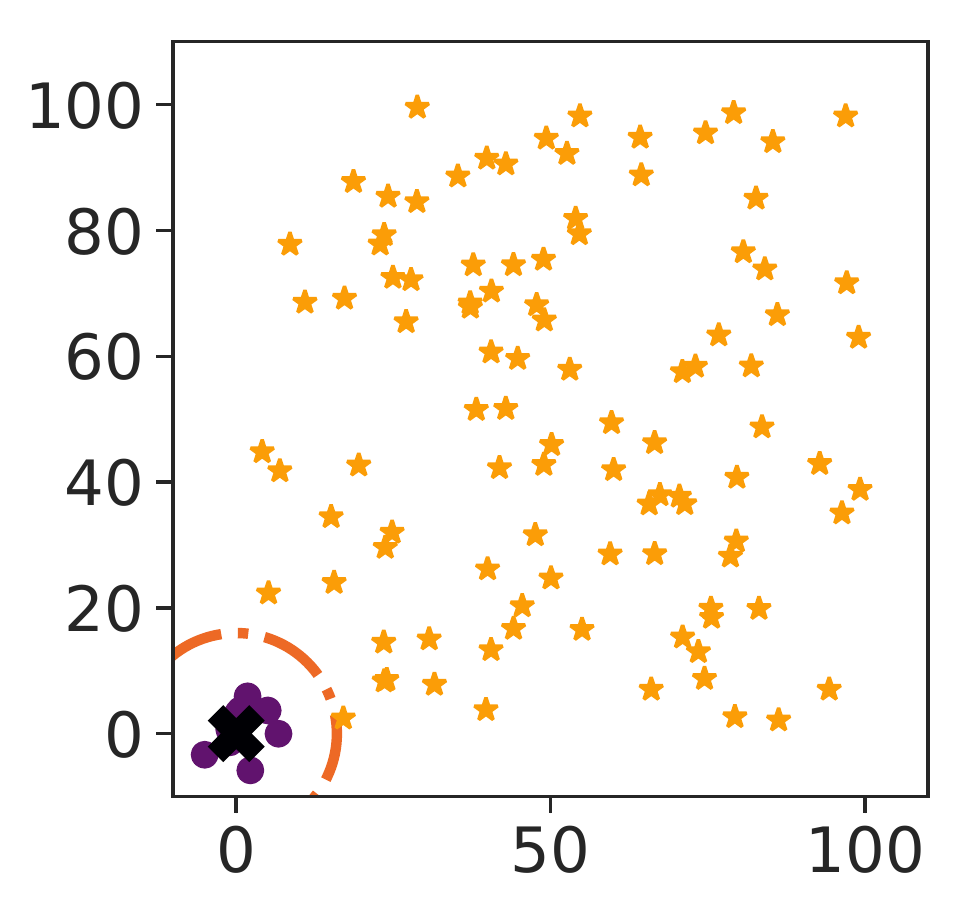}}
		\subcaption{10 seconds.}
		\label{fig_NA0p167-AtD12-M6-D1000_9p175}
	\end{subfigure}
	\hfill
	\begin{subfigure}[b]{0.22\textwidth}
		\centering
		\scalebox{1}[1]{\includegraphics[width=\textwidth]{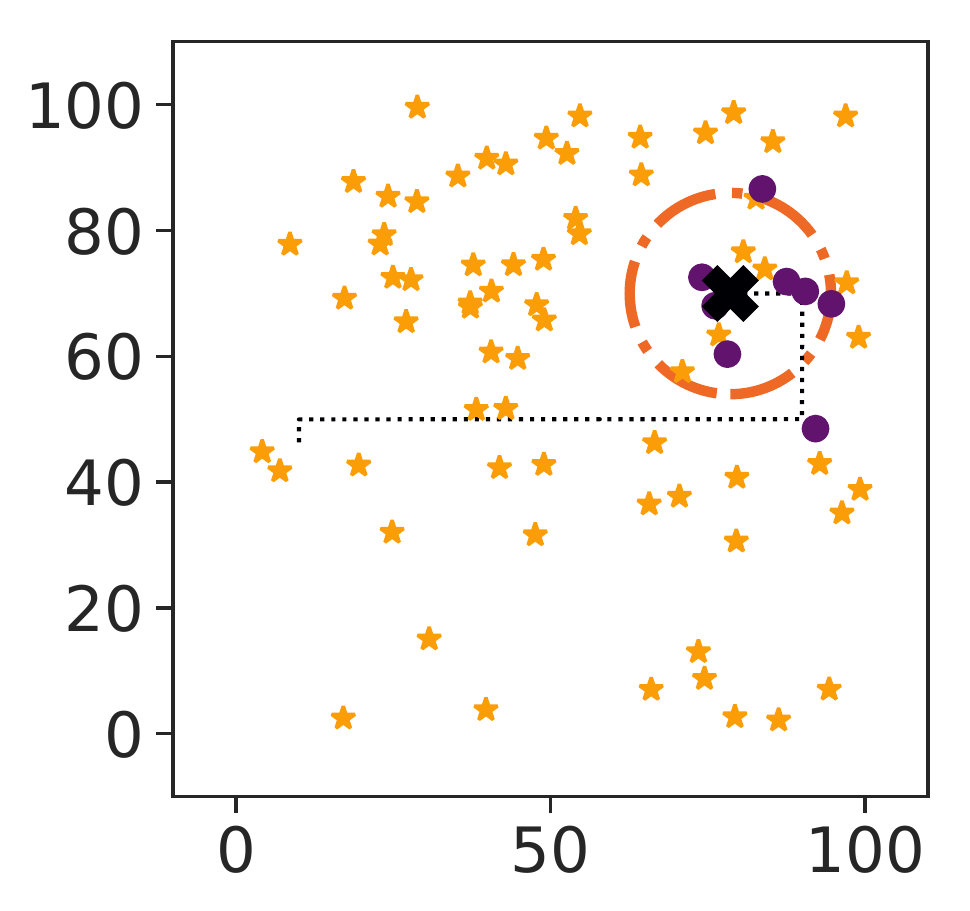}}
		\subcaption{3,500 seconds.}
		\label{fig_NA0p167-AtD12-M6-D1000_3367p175}
	\end{subfigure}\hfill
	\begin{subfigure}[b]{0.22\textwidth}
		\centering
		\scalebox{1}[1]{\includegraphics[width=\textwidth]{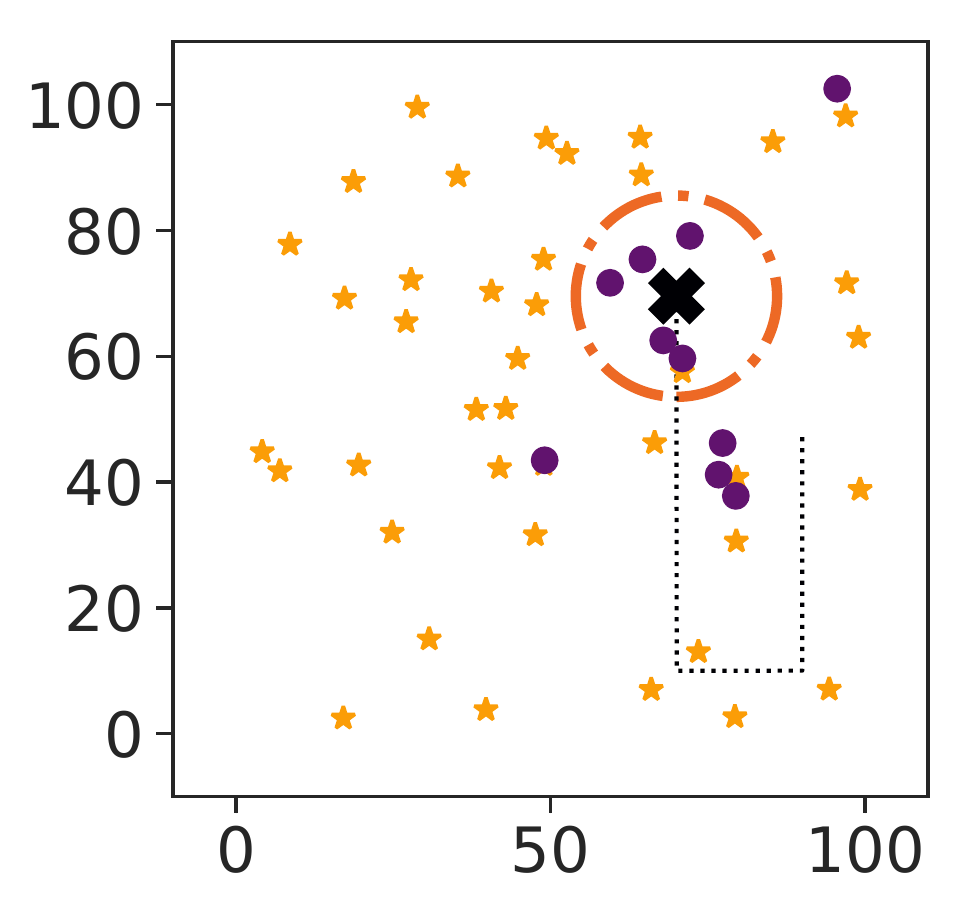}}
		\subcaption{7,000 seconds.}
		\label{fig_NA0p167-AtD12-M6-D1000_6725p175}
	\end{subfigure}\hfill
	\begin{subfigure}[b]{0.331\textwidth}
		\centering
		\scalebox{1}[1]{\includegraphics[width=\textwidth]{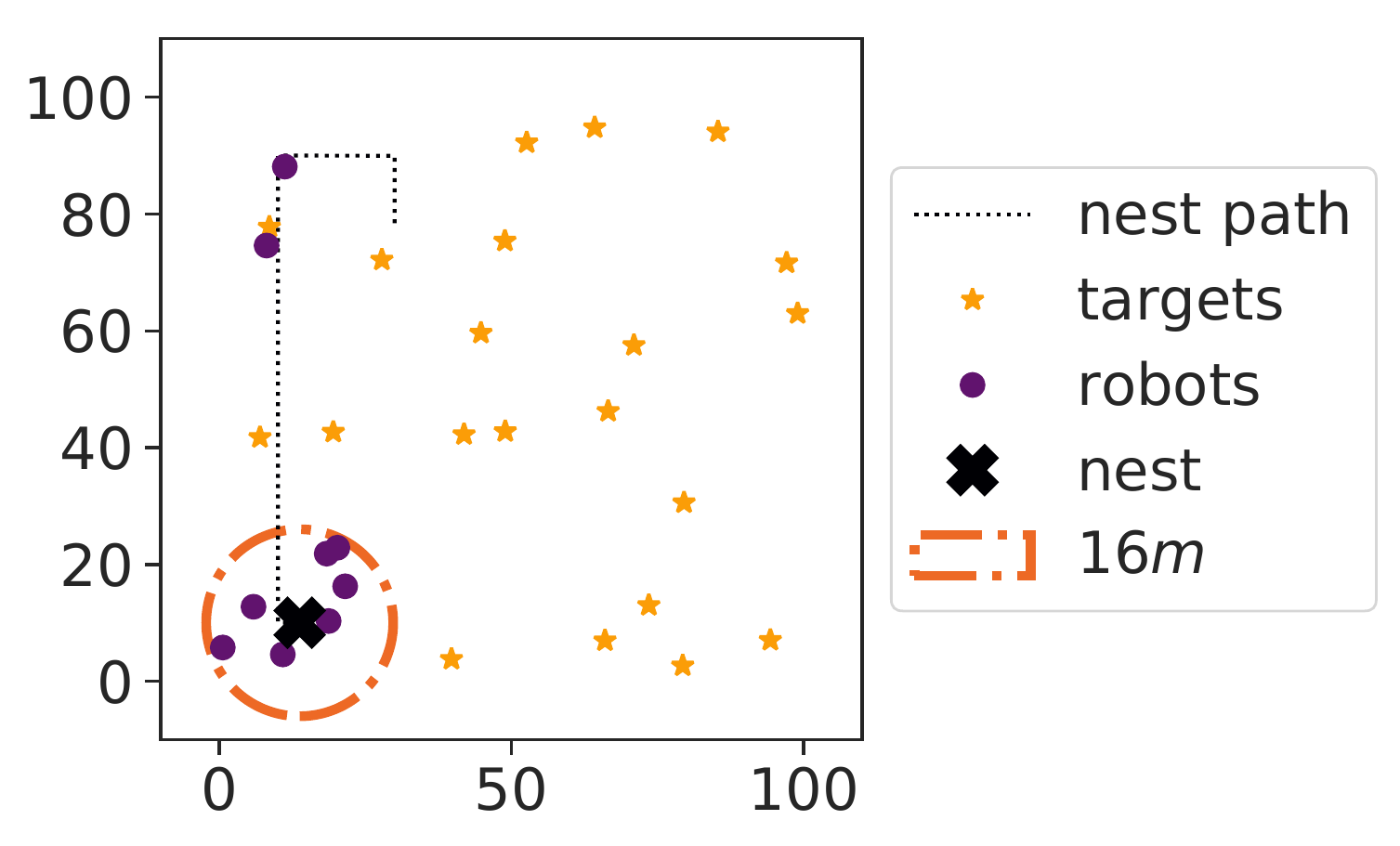}}
		\subcaption{10,000 seconds.}
		\label{fig_NA0p167-AtD12-M6-D1000_10092p375}
	\end{subfigure}\hspace{0.1\textwidth}\vspace{-.2cm}
	\caption{Sample simulation showing swarm searching for 100 targets uniformly distributed within 100m by 100m search area. (\ref{fig_NA0-AtD0-M1-D1_9p175} - \ref{fig_NA0-AtD0-M1-D1_10092p375}) are for $v_{n} = 0$ and robots perform random walk within a bounded search area. (\ref{fig_NA0p167-AtD12-M6-D1000_9p175} - \ref{fig_NA0p167-AtD12-M6-D1000_10092p375}) are for moving nest within unbounded search area, where $v_{n} = 0.167$, $M = 6$ and $D = 1000$. $x$ and $y$ axis are environment coordinates in metres.
	}
	\label{fig_moving_exploration_performance}\vspace{-.4cm}
\end{figure*}

In the stationary nest simulation setup, 100 targets were uniformly distributed within 14m radius search area around the nest. Number of found targets by the swarm at different time steps till 1000 simulated seconds, averaged for 30 independent simulation repetitions are shown in Table \ref{table_NA0_rad14_vs_unbounded}.
\begin{table}[t]
	\centering
	\caption{Comparison of number of targets found within a circular search area of radius 14m for $d_{c}$ of 10, 12 and 14 metres, compared to a swarm with with no chemotaxis behaviour in physically bounded and unbounded worlds. $M=10$ and $D=1000$ in all chemotaxis experiments.  Each value represents the mean and 95\% confidence interval based on 30 independent simulations.}\label{table_NA0_rad14_vs_unbounded}
	\resizebox{0.65\textwidth}{!}{\begin{tabular}{l@{\hskip 12pt}l@{\hskip 12pt}l@{\hskip 12pt}l@{\hskip 12pt}l@{\hskip 12pt}l}
			\toprule
			\diagbox[width=3em]{$\boldsymbol{t}$}{$\boldsymbol{d_{c}}$} &  Bounded & Unbounded & 10m & 12m & 14m \\
			\midrule
			200  &  $49.4 \pm 2.32$ &  $34.0 \pm 1.72$&     $49.2 \pm 1.32$ &     $46.7 \pm 1.43$ &     $45.3 \pm 1.12$ \\
			400  &  $71.2 \pm 2.27$ &  $39.0 \pm 2.18$&     $69.1 \pm 1.66$ &     $68.3 \pm 1.72$ &     $64.7 \pm 1.50$ \\
			600  &  $84.5 \pm 1.81$ &  $43.9 \pm 3.11$&     $79.3 \pm 1.33$ &     $81.1 \pm 1.40$ &     $77.0 \pm 1.55$ \\
			800  &  $91.2 \pm 1.25$ &  $47.0 \pm 3.43$&     $85.5 \pm 1.07$ &     $88.4 \pm 1.27$ &     $85.7 \pm 1.47$ \\
			1000 &  $95.2 \pm 0.97$ &  $50.5 \pm 3.79$&     $88.7 \pm 0.98$ &     $93.5 \pm 0.78$ &     $90.7 \pm 1.17$ \\
			\bottomrule
	\end{tabular}}
\end{table}
The result indicates that the exploration ability of the swarm when using chemotaxis to keep robots within the target search area is effective. Perfect restriction of the robot's movement to within the search area using a wall (\emph{Bounded}) caused the swarm to find 95.2\% of targets after 1000 seconds, while removal of the wall (and no chemotaxis) caused performance to drop to 50.5\% (\emph{Unbounded}). The presence of nest signal improved the swarm's performance in absence of a wall, causing them to locate 93.5\% when $d_{c} = 12$m. When $d_{c} = 10$m, the robots where able to effectively locate targets close to the nest at the early stages of the simulation, but the chemotactic region beyond 10m from the nest reduced the swarm's ability to locate targets in that region. Overall, $d_{c} = 12$m gave best balance between searching for targets within the chemotactic region (12m - 14m from nest) and the work area (0m - 12m), making it almost as good as the \emph{Bounded} case.

The last set of simulations investigates the swarm's ability to perform a similar exploration task within a 100m by 100m target search area, guided by a moving nest. As stated earlier, this is one of the main advantages of our approach, where swarms are able to follow a guide robot (or moving nest) while performing their tasks. In this setup, 100 targets were uniformly distributed within the world and the guide robot was used to perform a sweep of the environment using the search behaviour shown in Fig. \ref{fig_nest-navigation}. A value of $d_{c} = 12$m was used in all cases, while $v_{n}$ varied from 0.1 to 0.25. Simulations were stopped after 10,000 seconds. Table \ref{table_targets_foraged_vs_nest_vel} gives the average number of targets found by the swarm, while snapshots of sample simulations of the swarm performing the target search task are shown in Fig. \ref{fig_moving_exploration_performance}.

\begin{table}[t]
\centering	
	\caption{Targets found in a 100m by 100m world as exploration time, $t$, progresses. $d_{c} = 12$ metres, $M=6$ and $D=1000$ for the chemotaxis based approach. $v_{n} = 0$ for Bounded and Unbounded cases.
		Each value represents the mean and 95\% confidence interval based on 30 independent simulations.}\label{table_targets_foraged_vs_nest_vel}
	\resizebox{0.75\textwidth}{!}{\begin{tabular}{l@{\hskip 12pt}l@{\hskip 12pt}l@{\hskip 12pt}l@{\hskip 12pt}l@{\hskip 12pt}l@{\hskip 12pt}l}
			\toprule
			\diagbox[width=3em]{$\boldsymbol{t}$}{$\boldsymbol{v_{n}}$} & Bounded & Unbounded  &            0.1 &            0.125 &            0.167 &             0.25 \\
			\midrule
			2000  &  $35.1 \pm 1.21$&  $21.3 \pm 1.57$&      $17.9 \pm 0.84$ &        $19.3 \pm 1.29$ &        $22.9 \pm 1.30$ &       $21.9 \pm 1.45$ \\
			4000  &  $56.0 \pm 1.65$&  $27.0 \pm 2.02$&      $34.8 \pm 0.87$ &        $38.9 \pm 1.45$ &        $45.2 \pm 1.79$ &       $41.3 \pm 2.06$ \\
			6000  &  $69.0 \pm 1.69$&  $31.4 \pm 2.28$&      $51.7 \pm 1.11$ &        $57.9 \pm 1.60$ &        $61.9 \pm 2.03$ &       $55.2 \pm 2.18$ \\
			8000  &  $78.6 \pm 1.37$&  $34.3 \pm 2.45$&      $72.5 \pm 1.38$ &        $71.3 \pm 1.75$ &        $72.4 \pm 1.87$ &       $62.8 \pm 2.62$ \\
			10000 &  $86.1 \pm 1.30$&  $36.9 \pm 2.53$&      $80.8 \pm 1.09$ &        $78.4 \pm 1.61$ &        $79.1 \pm 1.53$ &       $70.9 \pm 2.68$ \\
			\bottomrule
		\end{tabular}
	}

\end{table}

After 10,000 simulation seconds, the swarm within a bounded search area was able to locate 86.1\% of targets (see Table \ref{table_targets_foraged_vs_nest_vel}). Removal of the wall caused swarm's performance to drop to 36.9\% of targets found. Using our chemotactic approach with a moving nest that guided exploration within the 100m by 100m search area, the swarm was able to give competitive target detection ability (80.1\% when $v_{n} = 0.1$). This is a significant contribution because it is a realistic approach to swarm robot deployment in open space (boundless) application areas, where it can be impractical to build fences around such regions or make the fence mobile to guide the swarm's work area.

In some applications (for example in foraging), it is important to find a balance between maximizing the number of robots close to the nest and fast exploration of the search area. A slow moving nest maximises robots close to the nest, resulting in thorough exploration of the work area. A fast moving nest, however, will result in the search area being covered quicker by the nest at the expense of the number of robots that are able to remain within the nest's work area. This causes the search to be less thorough. Multiple sweeps of the environment by the fast moving nest can compensate for this poor search.
We will investigate in more detail the question of maximising the number of robots close to the nest while minimising search time in future work.

\vspace{-.2cm}
\section{Real Robot Experiments}\label{sec_robot_experiments}
\vspace{-.2cm}
\begin{figure}[t]
	\centering
	\begin{subfigure}[b]{0.245\textwidth}
		\centering
		\scalebox{1}[1]{\includegraphics[width=\textwidth]{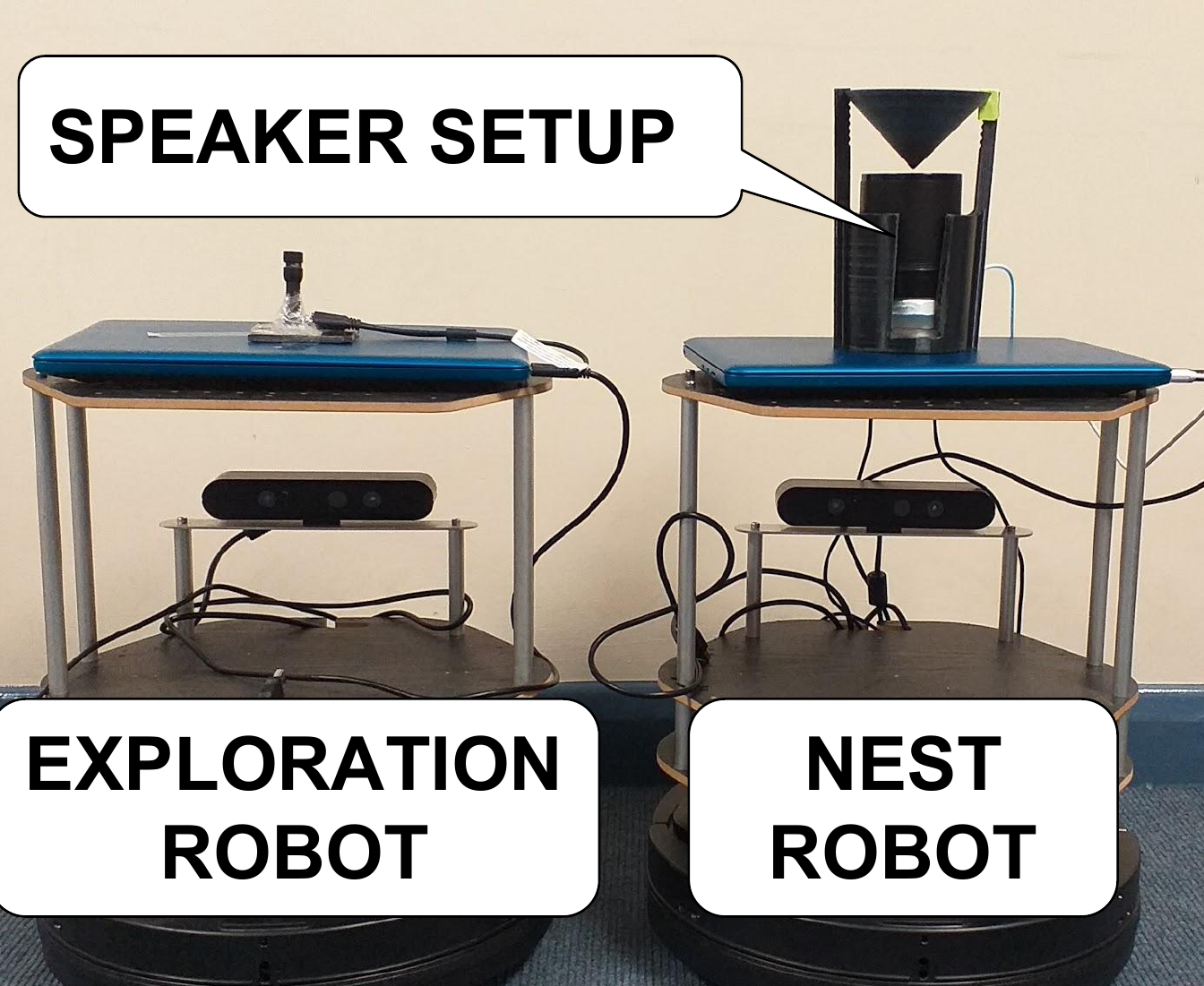}}
		\subcaption{Experiment robots.}
		\label{fig_turtlebot-setup}
	\end{subfigure}\hfill
	\begin{subfigure}[b]{0.74\textwidth}
		\centering	
		\scalebox{1}[1]{\includegraphics[width=\textwidth]{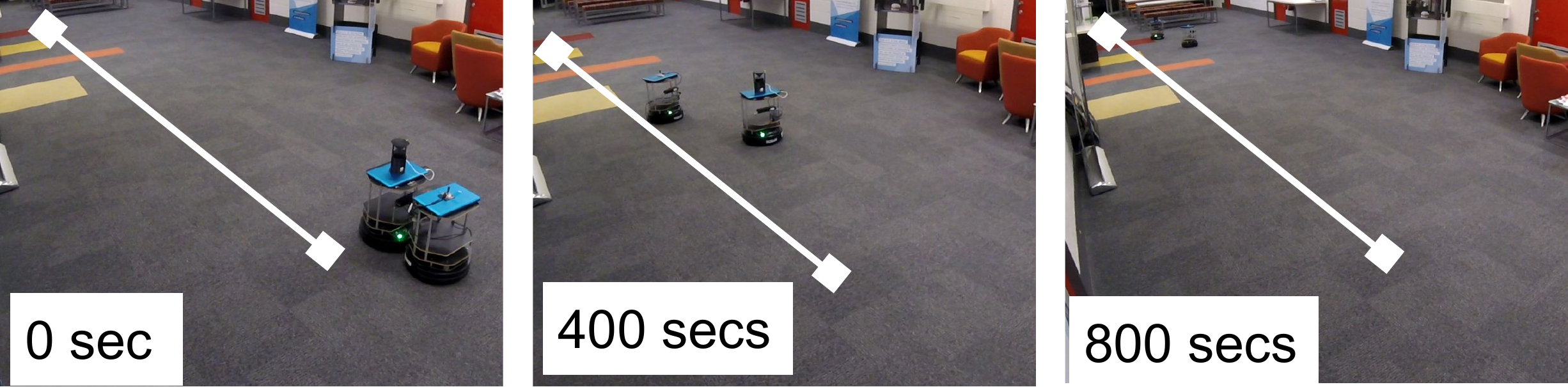}}
		\subcaption{White line is approximately 10 metres.}	
		\label{fig_follow-the-nest}
	\end{subfigure}\hfill
	\caption[]{Robots used for the experiment and snapshots of exploration robot following the moving nest. Experiment video is available at \url{https://youtu.be/ua0w3aXOYJI}.}
	\label{fig_turtlebot_snapshots}
	
	\vspace{-0.4cm}
\end{figure}

To validate that our simulation model can be realised on hardware, we conducted experiments with two Turtlebot2 robots\footnotemark\footnotetext{Two robots were used due to availability of robot hardware. Validation with more robots will be done in future work} where one robot acted as the nest and the second was used as the exploration robot. For this experiment, $v_{r} = 0.1$ m/s, $M = 6$, $D = 1000$ and $P_{b} = 0.0025$ per time step. The  nest robot used a speaker to broadcast white noise upwards, which was then reflected radially outwards using an inverted, 3D printed cone as shown in Fig. \ref{fig_turtlebot-setup}. The exploration robot used an omnidirectional microphone to perceive the sound signal. Work done by other researchers used multiple microphones on the robot to measure sound intensity  \cite{Arvin2018}. We show that our algorithm works well with a single microphone. Thereby simplifying the hardware implementation. Chemotaxis activation intensity, $A(d_{c}) = 180$ was used for these experiments.

\begin{figure}[t]
	\centering
	\begin{subfigure}[b]{0.4\textwidth}
		\centering
		\scalebox{1}[1]{\includegraphics[width=\textwidth]{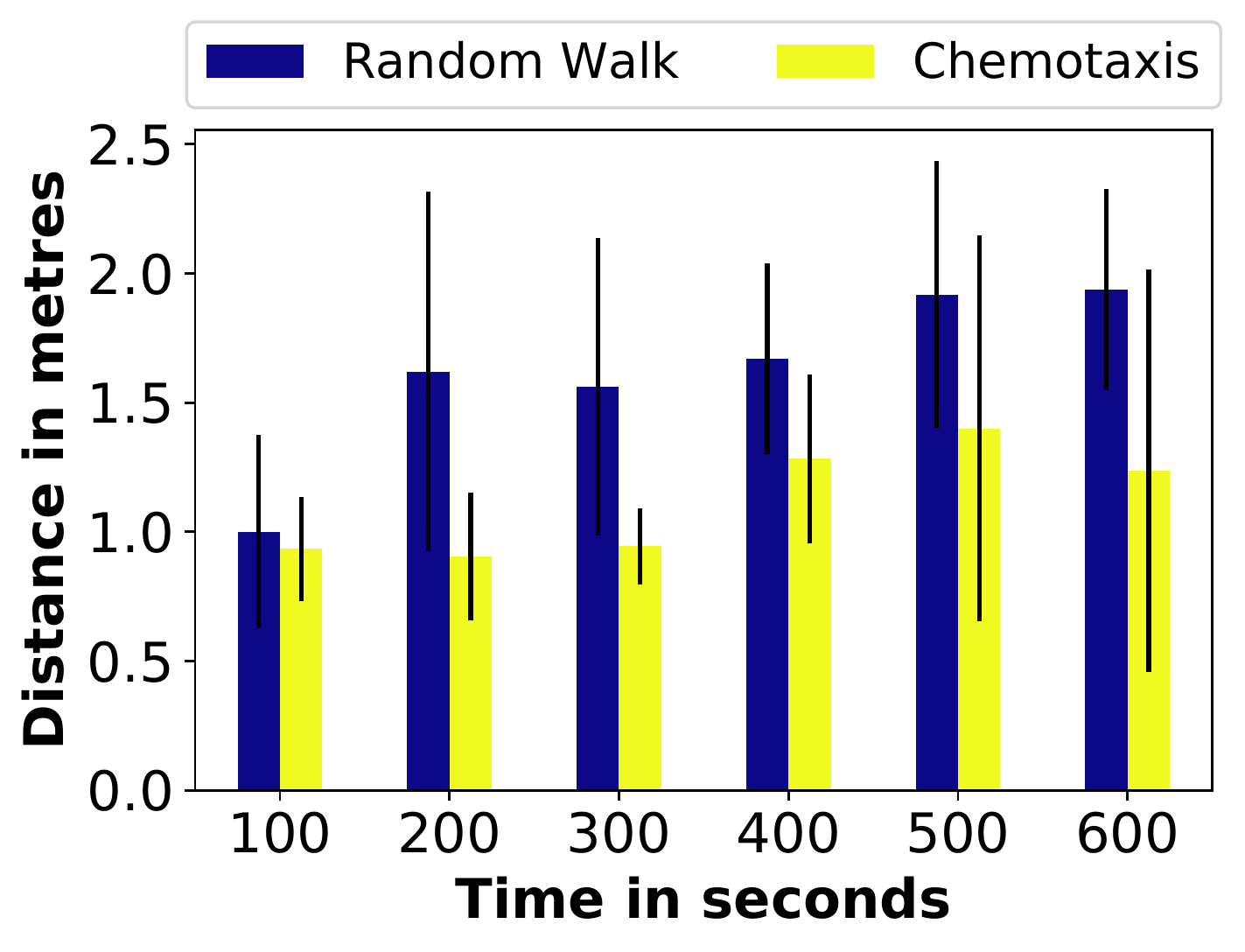}}
		\subcaption{Stationary nest. Distance was averaged for every 100 seconds duration.}
		\label{fig_stationaryci95-nest-following-turtlebot}
	\end{subfigure}\hfill
	\begin{subfigure}[b]{0.4\textwidth}
		\centering
		\scalebox{1}[1]{\includegraphics[width=\textwidth]{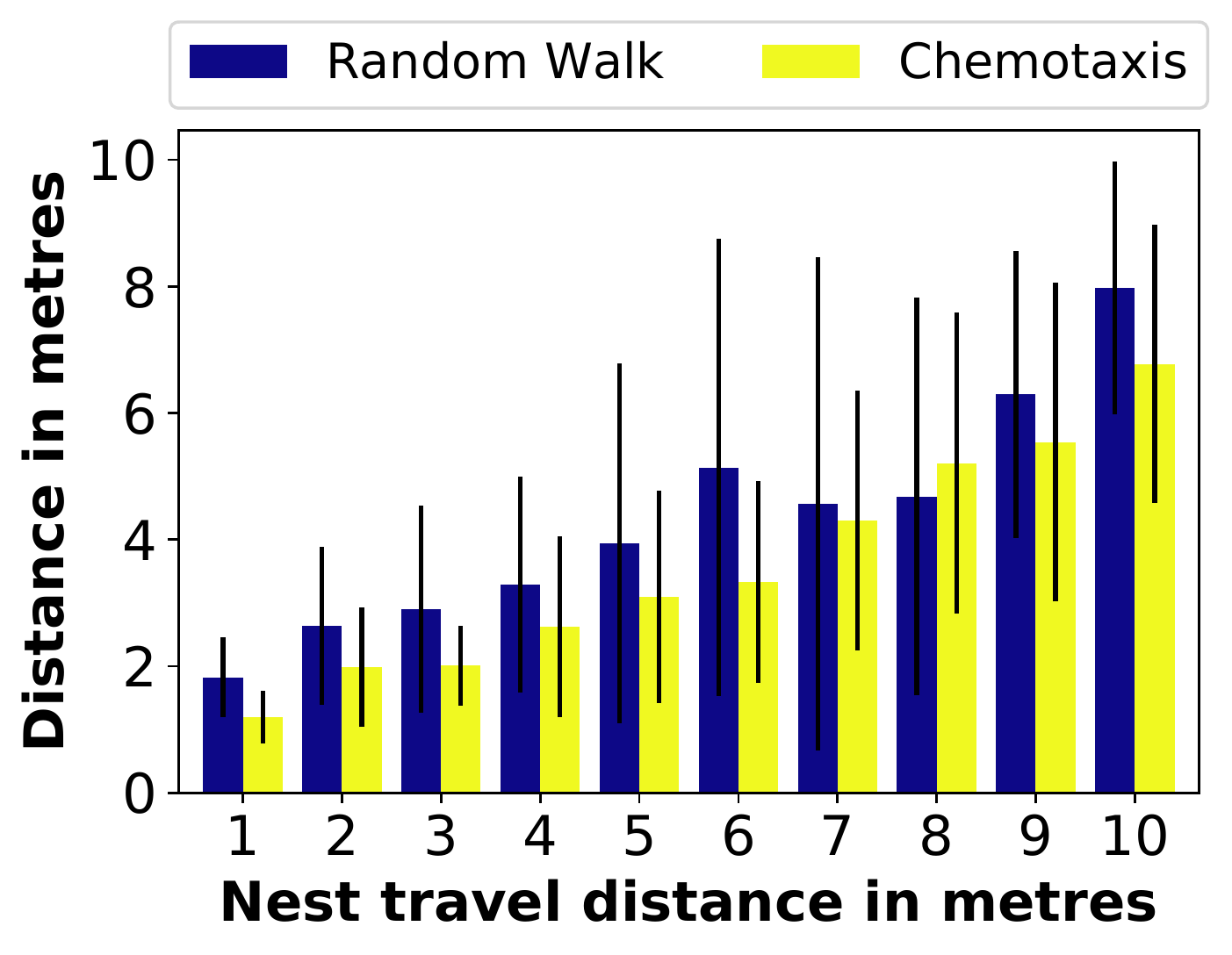}}
		\subcaption{Moving nest. Distance was averaged for every 1m of nest's journey.}
		\label{fig_20190318ci95-nest-following-turtlebot}
	\end{subfigure}
	\caption{Real robot validation of chemotaxis behaviour to remain close to nest. Experiments were repeated 5 times, and error bars represent 95\% confidence interval.}
	\label{fig_chemotaxis-validation-turtlebot}
	
	\vspace{-.4cm}
\end{figure}

For stationary nest experiments ($v_{n} = 0$) the nest was centred within a 3m by 6.4m space. The exploration robot was then left to perform random walk within the arena for 600 seconds. Fig. \ref{fig_stationaryci95-nest-following-turtlebot} compares the distance of the exploration robot from nest for random walk and our chemotaxis approach, averaged for 5 consecutive repetitions of the experiment. Results show that our approach was reasonably successful in keeping the exploration robot close to the nest's location, with mean distance being 0.9m within the first 100 seconds and 1.2m between 500 - 600 seconds of the experiment. This validates that the exploration robot was using the chemotaxis approach to remain close to the nest, and not just performing random exploration of the environment.

The second phase of the experiment measures the exploration robot's ability to remain close to a moving nest, when $v_{n} = 0.125$ and the distance travelled by nest is 10 metres. In this setup, the arena's width was 3m and one short edge of the arena was left open to allow the nest to make the 10m journey. Fig. \ref{fig_follow-the-nest} shows snapshots of the experiment as the nest makes its 10m journey. The average distance of exploration robot from the moving nest is shown in Fig. \ref{fig_20190318ci95-nest-following-turtlebot} for different distance ranges of the nest's journey. In comparison to the random walk, in which the exploration robot did not listen to the nest's signal, our chemotaxis approach indicates good nest following ability for the first 6m of the nest's journey. Beyond 6m, the chemotaxis became less effective. It is good to appreciate that echoes from the walls, intensity of sound source, ambient noise (including noise from the robots drive systems) and furniture in the environment can have a significant impact on the exploration robot's ability to compute a reliable temporal gradient when performing chemotaxis. These factors were the major contributors to poorer performance of our chemotaxis validation experiments compared to the simulations. In future research, we will work on minimising these environmental factors by conducting experiments in outdoor environments to eliminate effects of echoes and optimising the audio hardware for better signal-to-noise ratio.

\vspace{-.2cm}
\section{Conclusion}\label{sec_conclusion}
\vspace{-.2cm}
This paper has presented a simple, yet effective, means for deploying swarm robots in open (or boundless) environments. The biological inspiration for our algorithm is the chemotaxis behaviour used by the nematode \textit{C. elegans} to find high concentrations of chemical attractants.
We have used sound experiments to provide a realistic, hardware verifiable model of the communication used in our simulations. Extensive simulation experiments were conducted to investigate effects of our algorithm's parameters on the swarm's ability to use chemotaxis to return to the work area near the nest's location. Furthermore, we show that our algorithm is also effective in scenarios where the nest moves to guide the exploration robots to cover wider areas in a target search challenge, showing the interplay between nest velocity and number of robots that are able to keep up with it. Finally, we validated our algorithm using real robot experiments, showing that it is viable on hardware but would benefit from further optimisation.
In the future, we will extend the work to swarm foraging and investigate deployments on large scale swarm sizes on hardware platforms. We will also investigate other technologies, such as Wi-Fi, Zigbee and ultrasound for nest-robots communication.
\bibliographystyle{splncs04}
\bibliography{./samplepaper}
\end{document}